%% file: iclr2027_conference.tex
\documentclass{article} 
\usepackage{iclr2027_conference,times}

\input{math_commands.tex}

\usepackage{hyperref}
\usepackage{url}
\usepackage{graphicx}
\usepackage{booktabs} 
\usepackage{multirow}
\usepackage{wrapfig}
\usepackage{caption}
\usepackage[table]{xcolor}
\usepackage{amsmath,amssymb}
\usepackage{algorithm}
\usepackage{algpseudocode}

\title{Structural Process Supervision for Latent Chain-of-Thought Reasoning}

\author{Yiqi Li\textsuperscript{1}\quad Xu Chen\textsuperscript{2}\thanks{Corresponding author}\quad Chen Ju\textsuperscript{2}\quad Jiangchao Yao\textsuperscript{2}\quad Zhaoyang Li\textsuperscript{2}\quad Jinsong Lan\textsuperscript{2}\quad \\ \textbf{Xiaoyong Zhu\textsuperscript{2}\quad Yu Wang\textsuperscript{1}\footnotemark[1]}\\
\textsuperscript{1}Shanghai Jiao Tong University \quad
\textsuperscript{2}Taobao \& Tmall Group \\
\texttt{\{17-adamant, yuwangsjtu\}@sjtu.edu.cn}\\
\texttt{\{huaisong.cx\}@taobao.com}
}

\iclrfinalcopy 
\begin{document}

\maketitle

\begin{abstract}
Latent reasoning approaches enhance token-level efficiency and robustness by replacing verbose, explicit chain-of-thought (CoT) tokens with compact continuous-space embeddings. 
However, existing methods lack efficient process supervision over these embeddings, which often leads to representation collapse and uneven information distribution. 
To address this, we propose Prototype-Mediated Process Supervision (PMPS), which introduces learnable reasoning prototypes as semantic anchors to provide structural process-level supervision for latent reasoning. PMPS projects latent embeddings and explicit CoT embeddings into a shared prototype space, achieving many-to-many soft alignment between unequal-length representations through prototype assignment. Meanwhile, we introduce a Progressive Sequential Alignment (PSA) module to further guide training: positional priors initially encourage sequential alignment structure, then gradually relax to permit adaptive matching.
Experimental results show that PMPS compresses output token length to under 50\% of explicit CoT on GSM8K-Aug. Compared to leading baseline SIM-CoT, our method achieves average accuracy gains of 2.51\% across different model families. On GPT-2, accuracy of PMPS even surpasses CoT-SFT. On larger models and a more challenging task, PMPS consistently attains the highest accuracy among all latent reasoning methods with comparable output length.
\end{abstract}

\section{Introduction}
\begin{wrapfigure}{r}{0.48\textwidth}
  \vspace{-10pt}  
  \centering
  \includegraphics[width=0.48\textwidth]{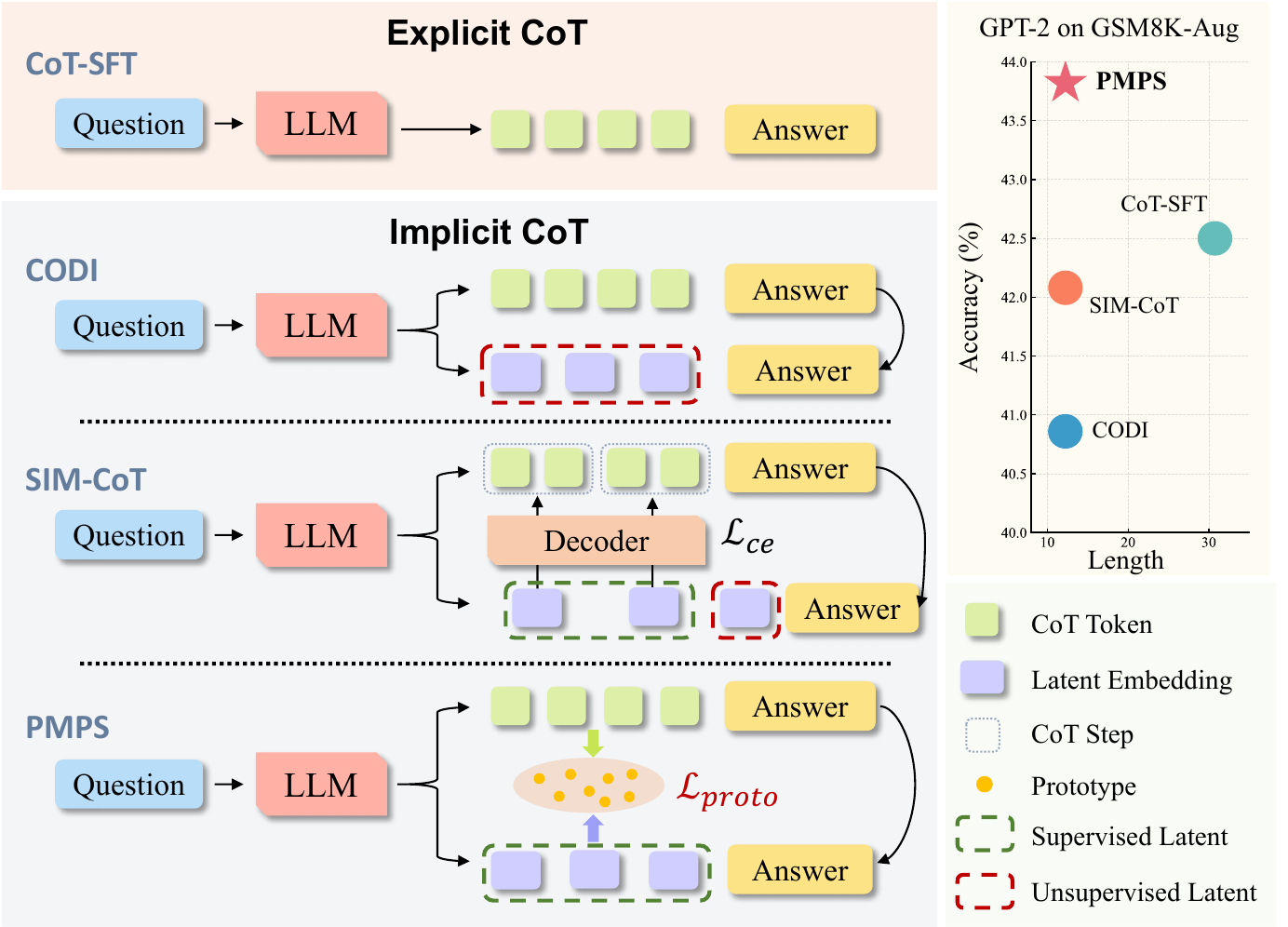}
  \caption{\textbf{Comparison of different CoT methods.} Our proposed PMPS provides process supervision over all latent embeddings.}
    \label{fig:fig1}
  \vspace{-10pt}  
\end{wrapfigure}
Large Language Models (LLMs) have demonstrated remarkable reasoning capabilities on complex tasks such as mathematical reasoning and logical inference \citep{achiam2023gpt,grattafiori2024llama,comanici2025gemini,yang2025qwen3,liu2025deepseek,zeng2026glm}, particularly when employing Chain-of-Thought (CoT) prompting techniques \citep{wei2022chain}. However, explicit CoT reasoning suffers from multiple limitations, including low inference efficiency and constrained expressiveness \citep{chen2025reasoning,li2025implicit,zhu2025surveylatentreasoning}, motivating researchers to explore more efficient and flexible reasoning paradigms.

To overcome the limitations of explicit CoT, researchers have proposed latent reasoning methods that perform reasoning in latent space rather than language space \citep{li2025latent,zhang2026soft,tan2026think}. Compared to explicit CoT, latent reasoning offers three advantages: \textbf{(1) Efficiency}: Replacing verbose CoT tokens with compact hidden states substantially reduces inference cost. \textbf{(2) Expressiveness}: Continuous high-dimensional representations transcend the discrete token bottleneck, encoding richer semantic information that natural language cannot faithfully capture\citep{deng2024explicit,hao2024coconut,deng2026llmlatentreasoningchain}. \textbf{(3) Controllability}: The reasoning length and implicit CoT are controllable by pre-setting the number of latent tokens.
Despite these advantages, existing latent reasoning approaches still suffer from critical limitations.
As shown in Figure \ref{fig:fig1}, CODI \citep{shen2025codi} optimizes latent embeddings solely via the final answer, which leaves intermediate reasoning unguided and often causes \emph{representation collapse}. SIM-CoT \citep{wei2025sim} mitigates this collapse via an auxiliary decoder to establish a one-to-one alignment between fixed-number latent embeddings and variable-length CoT steps. This design, which introduces a decoder of identical architecture to the base model, not only increases training overhead but also induces information bottlenecks, as excessive reasoning steps are compressed while surplus embeddings remain unsupervised. 
Empirical experiment (Figure \ref{fig:latent_nanlysis}) reveals two prevalent \emph{representation collapse} in existing methods: \textit{odd-even collapse}, where embeddings at odd and even positions collapse into two alternating groups; and \textit{tail collapse}, where the last few embeddings become nearly identical.

To address the above limitations, we propose \textbf{Prototype-Mediated Process Supervision (PMPS)}, a novel framework that provides structural process supervision for latent reasoning. 
We introduce learnable reasoning prototypes as semantic anchors and project both latent and explicit CoT embeddings into a unified prototype space, imposing structured semantic organization on the latent representations. Through Sinkhorn-Knopp \citep{cuturi2013sinkhorn} assignments, we establish soft many-to-many alignment between latent and CoT representations, and formulate a bidirectional cross-prediction loss that compels each latent embedding to capture differentiated reasoning semantics. 
Furthermore, we introduce a \textbf{Progressive Sequential Alignment (PSA)} module that applies a Gaussian positional prior to impose sequential structure on the matching, then gradually relaxes it via cosine annealing to allow adaptive alignment. Importantly, all components are used only during training, incurring only a small amount of training overhead and no additional inference cost.

Extensive experiments demonstrate the effectiveness of PMPS. On GSM8K-Aug, PMPS reduces the reasoning length by over 50\% relative to explicit CoT on GPT-2 while improving accuracy by 3.96\%, and it surpasses the CODI and SIM-CoT baselines by 4.33\% and 2.86\%, respectively. PMPS consistently achieves superior accuracy over all latent reasoning baselines across different model families, scales, and the more challenging task. In summary, our main contributions can be summarized as follows: 
\begin{itemize}
    \item We propose PMPS, the first framework to introduce structural process-level supervision for latent reasoning. PMPS establishes a soft alignment between unequal-length latent and explicit CoT embeddings via a prototype-based bidirectional cross-prediction objective, effectively mitigating \emph{representation collapse}.
    \item To preserve the intrinsic sequential reasoning structure of both CoT and latent embeddings, we introduce PSA, which employs a Gaussian positional prior with cosine annealing to guide early sequential alignment and gradually relaxes the constraint during training, enabling adaptive matching while balancing structural bias and flexibility.
    \item Extensive experiments across diverse benchmarks and model scales demonstrate that PMPS consistently outperforms leading latent reasoning baselines while preserving the inference efficiency, confirming both strong effectiveness and broad generalizability.
\end{itemize}

\section{Related Work}
\subsection{Latent Reasoning in LLMs} Chain-of-thought (CoT) prompting \citep{wei2022chain} improves LLM reasoning, but the inference cost increases substantially with verbose reasoning length, motivating reasoning in continuous hidden-state space.
iCoT \citep{deng2024explicit} and Coconut \citep{hao2024coconut} progressively replace explicit CoT tokens with latent embeddings via curriculum learning. CODI \citep{shen2025codi} stabilizes training through teacher–student distillation at the answer position. SIM-CoT \citep{wei2025sim} enforces rigid one-to-one assignment between CoT steps and latent embeddings with an auxiliary decoder. CoLaR \citep{tan2026think} and Latent-SFT \citep{deng2026llmlatentreasoningchain} compress a contiguous sequence of CoT tokens into a single latent embedding. However, these methods either lack direct reasoning supervision or enforce an overly rigid correspondence between the CoT and latent embeddings, which limits the expressiveness of latent representations. We propose PMPS to address these limitations via a structural prototype-mediated soft alignment between the CoT and latent embeddings in latent space, requiring \textit{no auxiliary decoder} and \textit{no hard CoT partition}.

\subsection{Process Supervision and Prototype-Based Representation Learning} Process Reward Models \citep{lightman2024let} show that step-level correctness labels yield better reasoning verifiers than outcome-only rewards, with follow-up work automating such supervision via Monte Carlo tree search \citep{wang2024math}. However, existing process supervision methods require explicit, discrete reasoning steps, which are incompatible with continuous latent embeddings. In parallel, SwAV \citep{caron2020unsupervised} and PCL \citep{li2020prototypical}, originally from visual self-supervised learning, show that learnable prototypes can effectively structure embedding spaces without negative pairs. Knowledge distillation approaches \citep{sun2019patient,wang2020minilm} align teacher–student representations but typically assume matched sequence lengths. Our work bridges these by leveraging the prototype-based assignment and cross-prediction objective to provide many-to-many structural supervision across the unequal number of latent and CoT embeddings, complemented by answer-position distillation from the teacher to ensure outcome-level alignment.
\begin{figure}[t]
    \centering
    \includegraphics[width=0.9\linewidth]{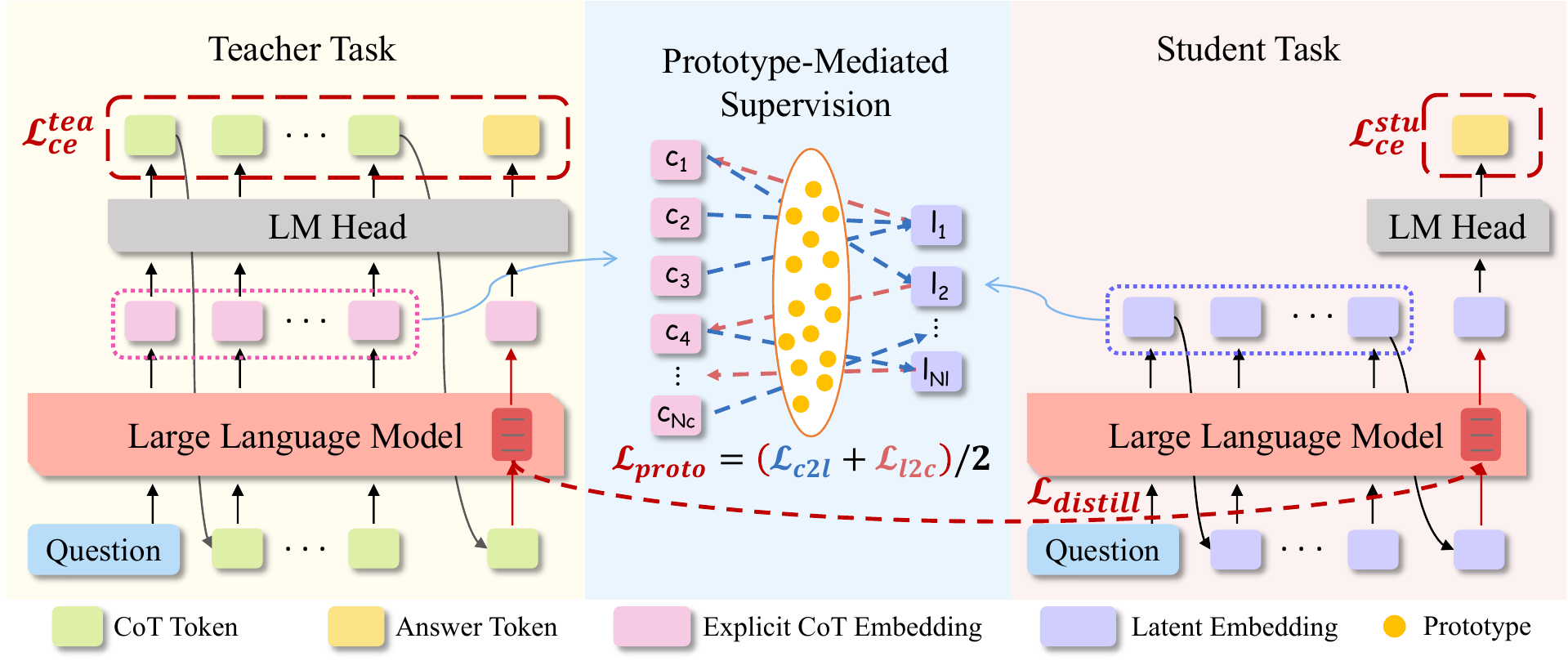}
    \caption{\textbf{Overview of PMPS framework.} During training, PMPS adopts the conventional parallel teacher-student self-distillation architecture and leverages the explicit CoT embeddings from the teacher task to supervise the latent embeddings from the student task via a bidirectional cross-prediction loss in the prototype space. At inference, only the student task is executed, performing latent reasoning without any additional overhead.}
    \label{fig:method}
\end{figure}
\vspace{-10pt}
\section{Methodology} \label{sec:method}
In this section, we present the technical details of our framework. We first review the conventional self-distillation framework. We then introduce PMPS, which provides structural supervision for latent embeddings. Finally, we describe the PSA module, which guides sequential alignment structure during training. The overview of our framework is presented in Figure \ref{fig:method} and Appendix \ref{sec:psedo}.

\subsection{Preliminaries}\label{sec:prelim}
To obtain high-quality explicit CoT embeddings that align with the model's intrinsic distribution as supervision signals, we adopt the widely used self-distillation framework \citep{zhang2019your,liao2023slsd,liu2026ssr} as the foundational architecture. Within this framework, the model is jointly optimized for two tasks: explicit CoT reasoning (teacher task) and latent reasoning (student task). Inspired by CODI \citep{shen2025codi}, we align the hidden states of teacher and student task at the answer token positions. The objective function for this base architecture is formulated as:
\begin{equation}
    \mathcal{L}_{\text{base}} = \mathcal{L}_{ce}^{stu} + \mathcal{L}_{ce}^{tea} + \lambda_{d} \cdot \mathcal{L}_{dist}
\end{equation}
where $\mathcal{L}_{ce}^{stu}$ and $\mathcal{L}_{ce}^{tea}$ are cross-entropy losses for the student's answer prediction and the teacher's full-sequence prediction, $\mathcal{L}_{dist}$ is a Smooth L1 distillation loss aligning teacher and student hidden states at the answer token, and $\lambda_{d}$ is the weighting coefficient for distillation loss. However, relying solely on this base architecture lacks process-level supervision for the latent embeddings, which can lead to latent homogenization and representation collapse.

\subsection{Prototype-Mediated Process Supervision}
To mitigate the representation collapse induced by the absence of process supervision, we propose to leverage the CoT embeddings from the teacher task as supervision signals. These explicit CoT embeddings are well-suited targets for two reasons: (1) the $\mathcal{L}_{ce}^{tea}$ loss ensures that they encode high-quality, semantically meaningful reasoning information, and (2) since both the explicit CoT embeddings and latent embeddings are generated by the same model, they share the same hidden-state space, providing a common representational basis for alignment.

However, direct one-to-one distillation is infeasible because the number of latent embeddings $N_l$ is far smaller than the number of CoT tokens $N_c$. 
We therefore propose Prototype-Mediated Process Supervision (PMPS), which introduces $K$ learnable prototype vectors $\mathbf{P} = \{\mathbf{p}_1, \ldots, \mathbf{p}_K\} \in \mathbb{R}^{K \times d}$ as shared semantic anchors to mediate a soft many-to-many alignment between the two sets of representations, imposing structured semantic organization on the latent space. 
Let $\mathbf{H}_{l} = \{\mathbf{l}_1, \ldots, \mathbf{l}_{N_l}\} \in \mathbb{R}^{N_l \times D}$ and $\mathbf{H}_{c} = \{\mathbf{c}_1, \ldots, \mathbf{c}_{N_c}\} \in \mathbb{R}^{N_c \times D}$ denote the latent embeddings and the explicit CoT embeddings, respectively, where $D$ is the hidden dimension and $d$ is the prototype dimension.

\textbf{Projection and Scoring.}
A shared two-layer MLP with batch normalization serves as the projection head $g(\cdot)$, projecting both $\mathbf{H}_{l}$ and $\mathbf{H}_{c}$ from the original hidden space into the prototype space, followed by $\ell_2$ normalization:
\begin{equation}
    \mathbf{z} = g(\mathbf{h}) / \| g(\mathbf{h}) \|_2
\end{equation}
where $\mathbf{h}$ denotes either the latent embedding or the explicit CoT embedding. Each unit feature vector $\mathbf{z}$ is scored against the $\ell_2$-normalized prototypes via inner product $s_k = \mathbf{z}^\top \mathbf{p}_k$, yielding score matrices $\mathbf{S}_{l} \in \mathbb{R}^{N_l \times K}$ and $\mathbf{S}_{c} \in \mathbb{R}^{N_c \times K}$.

\textbf{Prototype Assignment.}
To obtain soft assignments while preventing prototype collapse, we apply the Sinkhorn-Knopp algorithm \citep{cuturi2013sinkhorn} to enforce an equipartition constraint. Given a score matrix $\mathbf{S} \in \mathbb{R}^{M \times K}$, where $M$ denotes the number of tokens, we initialize $\mathbf{Q}^{(0)} \propto \exp(\mathbf{S}/\epsilon)$, with $\epsilon$ serving as the regularization parameter. We then iteratively alternate row and column normalization:
\begin{equation}
    \mathbf{Q}^{(t)} \leftarrow \text{Norm}_{\text{row}}\!\left(\text{Norm}_{\text{col}}\!\left(\mathbf{Q}^{(t-1)}\right)\right)
\end{equation}
The converged $\mathbf{Q} \in \mathbb{R}^{M \times K}$ nearly satisfies that each row sums to $1$ (a valid per-token distribution) and each column sums to $M/K$ (uniform prototype utilization) after three iterations. This procedure is applied independently to both $\mathbf{S}_{l}$ and $\mathbf{S}_{c}$ with stop-gradient operation following SwAV \citep{caron2020unsupervised}, producing the final soft assignment codes $\mathbf{Q}_{l} \in \mathbb{R}^{N_l \times K}$ and $\mathbf{Q}_{c} \in \mathbb{R}^{N_c \times K}$. Detailed Sinkhorn-Knopp algorithm is show in Appendix \ref{sec:sink}.

\textbf{Many-to-Many Matching.}
Leveraging the prototype assignments as a bridge, we compute the soft matching matrix between latent embeddings and explicit CoT embeddings:
\begin{equation}
    \mathbf{M}_{c2l} = \mathbf{Q}_{c} \cdot \mathbf{Q}_{l}^\top / \tau_m \in \mathbb{R}^{N_c \times N_l}
\end{equation}
where temperature $\tau_m$ controls matching sharpness. We then apply softmax normalization along the last dimensions to obtain bidirectional matching weights:
\begin{equation}
    \mathbf{A}_{c2l} = \text{softmax}(\mathbf{M}_{c2l}), \quad \mathbf{A}_{l2c} = \text{softmax}(\mathbf{M}_{c2l}^\top)
\end{equation}
This formulation naturally induces a many-to-many alignment: a single latent embedding can attend to multiple explicit CoT embeddings for compression, while a single explicit CoT embedding can contribute to multiple latent embeddings for semantic sharing.

\textbf{Bidirectional Cross-Prediction Loss.}
Alongside the stop-gradient assignment codes $\mathbf{Q}$, we compute gradient-carrying prototype predictions $\boldsymbol{\phi}_{l} = \text{softmax}(\mathbf{S}_{l}/\tau)$ and $\boldsymbol{\phi}_{c} = \text{softmax}(\mathbf{S}_{c}/\tau)$, where $\tau$ is a prediction temperature. For each explicit CoT embedding $j$, we aggregate the matched latent predictions weighted by $\mathbf{A}_{c2l}$ and minimize the cross-entropy with its assignment code:
\begin{equation}
    \mathcal{L}_{c2l} = -\frac{1}{N_c} \sum_{j=1}^{N_c} \sum_{k=1}^{K} Q_{c}^{(j,k)} \log \left(\sum_{i=1}^{N_l} A_{c2l}^{(j,i)} \cdot \phi_{l,i}^{(k)}\right)
\end{equation}
where $Q_{c}^{(j,k)}$ is the $(j,k)$-th entry of $\mathbf{Q}_{c}$, $A_{c2l}^{(j,i)}$ is the matching weight from the $j$-th explicit CoT embedding to the $i$-th latent embedding, and $\phi_{l,i}^{(k)}$ is the predicted probability of the $i$-th latent embedding for the $k$-th prototype. Symmetrically, the latent-to-CoT direction yields:
\begin{equation}
    \mathcal{L}_{l2c} = -\frac{1}{N_l} \sum_{i=1}^{N_l} \sum_{k=1}^{K} Q_{l}^{(i,k)} \log \left(\sum_{j=1}^{N_c} A_{l2c}^{(i,j)} \cdot \phi_{c,j}^{(k)}\right)
\end{equation}
The overall process supervision loss is $\mathcal{L}_{proto} = \frac{1}{2}(\mathcal{L}_{c2l} + \mathcal{L}_{l2c})$. The CoT-to-latent direction ($\mathcal{L}_{c2l}$) ensures that the latent embeddings collectively explain each explicit CoT embedding, while the latent-to-CoT direction ($\mathcal{L}_{l2c}$) encourages each latent embedding to capture diverse CoT semantics.

\subsection{Progressive Sequential Alignment}\label{sec:psa}
Explicit CoT inherently follows a progressive process: early tokens build up the reasoning chain, while final tokens converge toward the answer. However, the unconstrained matching in the base PMPS allows any latent token to align with any CoT position, which risks failing to capture the progressive reasoning structure. Conversely, enforcing a rigid segmented mapping is overly restrictive: uniform partitioning introduces noisy supervision that undermines the flexibility of semantic-driven matching. To navigate between these two extremes, we introduce Progressive Sequential Alignment (PSA), which injects sequential structure into the matching through a soft Gaussian positional prior and gradually relaxes it via curriculum-scheduled decay.
Concretely, we augment $\mathbf{M}_{c2l}$ with a bias penalizing positionally distant matches:
\begin{equation}
    \hat{M}_{c2l}^{(j,i)} = M_{c2l}^{(j,i)} - \alpha \cdot \left(\frac{j-1}{N_c - 1} - \frac{i-1}{N_l - 1}\right)^2
    \label{eq:psa}
\end{equation}
where ($u_j=(j-1)/(N_c-1)$) and ($v_i=(i-1)/(N_l-1)$) are normalized positions, and ($\alpha$) controls the prior strength. This bias acts as a multiplicative Gaussian prior ($\exp[-\alpha(u_j-v_i)^2]$), favoring matches at similar relative positions \citep{Su_2017_CVPR,yang-etal-2018-modeling}. As the latent index increases, the preferred CoT region shifts forward, encouraging sequential alignment while retaining semantic flexibility without enforcing strict monotonicity. The biased scores replace ($M_{c2l}$) in both directions of the matching normalization.
To balance structural guidance with representational flexibility, we adopt cosine annealing to decay $\alpha$ from $\alpha_{\max}$ to $0$ over training:
\begin{equation}
    \alpha(t) = \frac{\alpha_{\max}}{2} \left(1 + \cos\left(\pi \cdot \frac{t}{T}\right)\right)
\end{equation}
where $t$ and $T$ are the current and total training steps. Early in training ($\alpha$ large), the strong prior helps latent embeddings establish sequential structure and stabilizes matching. Late in training ($\alpha \to 0$), the constraint is fully released, allowing adaptive alignment where reasoning steps may span or consolidate across latent tokens. PSA modifies only the matching scores before softmax, introducing no learnable parameters and incurring zero inference overhead.

\subsection{Overall Training Objective}
The overall training objective combines the base loss with our proposed process supervision loss:
\begin{equation}
    \mathcal{L}_{total} = \mathcal{L}_{base} + \lambda_{proto} \cdot \mathcal{L}_{proto}
\end{equation}
where $\lambda_{proto}$ is the weighting coefficient for the process supervision loss. 

The teacher task, prototype layer, projection head, and PSA module are used exclusively during training. 
At inference time, the model architecture and computational cost are identical to the student task. This design ensures that all accuracy improvements come at zero additional inference cost.

\begin{table}[ht]
\caption{\textbf{Performance comparison on in-domain dataset and out-of-domain datasets.} Experiments are conducted on three different backbones, reporting answer accuracy (Acc.) and output token length (Len.). Explicit CoT method is marked with \colorbox{gray!20}{\phantom{xxx}}. The best and second best accuracy among latent reasoning approaches are highlighted in \textbf{bold} and \underline{underline}, respectively.}
\label{tab:main}
\centering
\resizebox{\linewidth}{!}{
\begin{tabular}{lccccccccc}
\toprule
\multicolumn{1}{c}{\multirow{3}{*}{\textbf{Method}}} & \multicolumn{2}{c}{\textbf{In-domain}} & \multicolumn{5}{c}{\textbf{Out-of-domain}}& \multicolumn{2}{c}{\multirow{2}{*}{\textbf{Average}}} \\ \cmidrule(rl){2-3} \cmidrule(rl){4-8}
 & \multicolumn{2}{c}{GSM8K-Aug} & \multicolumn{1}{c}{GSM-Hard} & \multicolumn{1}{c}{SVAMP} & \multicolumn{1}{c}{MultiArith} & \multicolumn{2}{c}{Average} &\\ 
\cmidrule(rl){2-3} \cmidrule(rl){4-4} \cmidrule(rl){5-5} \cmidrule(rl){6-6} \cmidrule(rl){7-8} \cmidrule(rl){9-10}
 & Acc. (\%) & Len. & Acc. (\%)  & Acc. (\%)& Acc. (\%)& Acc. (\%)& Len.& Acc. (\%)& Len. \\ 
\cmidrule(rl){1-10}
\multicolumn{10}{c}{\textit{GPT-2}} \\ 
\cmidrule(rl){1-10}
\rowcolor{gray!20} CoT-SFT  & 42.50 & 30.76 & 9.25 & 40.00  & 92.22  &47.16 &25.08 & 45.99 & 26.50 \\
Coconut & 36.50 & 15.40 & 7.58 & 34.67 & 82.20 &41.48 &11.87 & 40.24& 12.75 \\
CoLaR-2 & 11.07 & 14.43 & 2.58  & 15.00 & 41.11 & 19.56 &10.89& 17.44 & 11.78 \\
CODI & 40.86 & 12.23 & \underline{9.40} & 40.00   & 92.22  & 47.21& 12.89&45.62& 12.72 \\
SIM-CoT & \underline{42.08} & 12.24 & \underline{9.40} & \underline{43.00}& \underline{93.89}& \underline{48.76} & 12.58&\underline{47.09}  & 12.49 \\
PMPS(Ours) & \textbf{44.35} & 12.24& \textbf{10.46}& \textbf{48.33}& \textbf{96.67} &\textbf{51.82} & 12.58&\textbf{49.95(+2.86)}& 12.49 \\ 
\cmidrule(rl){1-10}
\multicolumn{10}{c}{\textit{Qwen2.5-0.5B-Instruct}}     \\  
\cmidrule(rl){1-10}
\rowcolor{gray!20} CoT-SFT &59.82 &42.68 & 16.91 & 61.67  & 97.78 & 58.79 &38.91& 59.05 & 39.85\\
Coconut &15.69 & 12.53 & 3.26 & 26.33 & 22.78 & 17.46 & 13.01 & 17.02 & 12.89\\
CoLaR-2 &29.57 & 17.92 & 6.44 & 37.00& 77.78 & 40.41 & 16.04 & 37.70 & 16.51\\
Latent-SFT &\underline{43.21} & 13.09 & 9.86 & 50.66 & 91.67 &50.73 & 15.60 & 48.85 & 14.97    \\
CODI &39.35 & 14.43 & 9.33 & 48.33 & 83.89 & 47.18 & 15.03 & 45.23 & 14.88\\
SIM-CoT &42.08 & 14.45 & \underline{10.16} & \underline{51.67} & \underline{92.22} & \underline{51.35} & 15.03 & \underline{49.03} & 14.91\\
PMPS(Ours) & \textbf{46.70} & 14.43 & \textbf{10.84} & \textbf{57.33} & \textbf{95.00} & \textbf{54.39} & 15.68 & \textbf{52.47(+3.44)} & 15.37 \\
\cmidrule(rl){1-10}
\multicolumn{10}{c}{\textit{Llama-3.2-1B-Instruct}}   \\                             
\cmidrule(rl){1-10}
\rowcolor{gray!20} CoT-SFT & 63.00& 32.51& 15.09& 66.33& 98.33 & 59.92 & 27.07 & 60.69 & 28.43\\
Coconut & 40.03 & 16.44 & 8.42 & 47.33 & 83.89 & 46.55 & 13.80 & 44.92 & 14.46 \\
CoLaR-2 & 42.15 & 12.70 & 8.95 & 55.33 & 91.67 & 51.98 & 13.41 & 49.53 & 13.24  \\
Latent-SFT & 48.82 & 12.83 & 10.54 & 50.33 & 90.56 & 50.48 & 9.65 & 50.06 & 10.44 \\
CODI & 53.90 & 13.21 & 12.28 & 56.00 & \textbf{97.78} &55.35 & 11.46 & 54.99 & 11.90 \\
SIM-CoT & \underline{56.03} & 13.20 & \underline{12.36} & \underline{60.33} & \underline{97.22} & \underline{56.64} & 13.47 &\underline{56.49} & 13.41 \\
PMPS(Ours) & \textbf{56.56} & 13.19 & \textbf{13.42} & \textbf{63.67} & \underline{97.22} & \textbf{58.10} & 13.51&\textbf{57.72(+1.23)} & 13.43 \\ 
\bottomrule
\end{tabular}
}
\vspace{-8pt}
\end{table}

\section{Experiments}
\subsection{Experimental Setup} \label{sec:setup}
\paragraph{Training and Evaluation Data.} Following prior work on latent reasoning~\citep{shen2025codi,wei2025sim}, we primarily train and evaluate on \textbf{GSM8K-Aug}~\citep{deng2023gsm8k-aug}, an augmented version of the GSM8K~\citep{cobbe2021gsm8k} where CoT annotations are structured arithmetic expressions rather than natural language.
To assess the generalization ability of our method, we additionally evaluate on three out-of-domain (OOD) benchmarks: (1) \textbf{GSM-Hard} \citep{gao2023gsm-hard}, (2) \textbf{SVAMP} \citep{patel2021svamp}, (3) \textbf{MultiArith} \citep{roy2015multiarith}. Detailed information is provided in Appendix \ref{sec:bench}.
We adopt two evaluation metrics: \textbf{Accuracy (Acc.)} measures the correctness of the final numerical answer, and \textbf{Length (Len.)} measures the total number of generated tokens.

\paragraph{Baselines.} We compare our approach against six representative baselines across the following four categories: (1) Explicit CoT methods: \textbf{CoT-SFT}; (2) Curriculum learning-based methods: \textbf{Coconut}~\citep{hao2024coconut}; (3) Token compression-based methods: \textbf{CoLaR-2}~\citep{tan2026think} and \textbf{Latent-SFT}~\citep{deng2026llmlatentreasoningchain}; (4) Self-distillation-based methods: \textbf{CODI}~\citep{shen2025codi} and \textbf{SIM-CoT}~\citep{wei2025sim}. 
Notably, the latter three categories constitute latent reasoning methods. Detailed descriptions of all baselines are provided in Appendix \ref{sec:baseline}.

\paragraph{Implementation Details.} We conduct experiments on three model families: GPT-2~\citep{radford2019gpt2}, Qwen2.5-0.5B-Instruct~\citep{yang2024qwen2.5}, and LLaMA-3.2-1B-Instruct~\citep{grattafiori2024llama}. For a fair comparison, all methods except Coconut adopt LoRA~\citep{hu2022lora} fine-tuning with the same rank of 128; Coconut uses full-parameter fine-tuning as in its original implementation. For SIM-CoT, we adopt the CODI-based variant for a direct comparison under the same backbone. All other training hyperparameters follow the respective original papers and formal repositories. For PMPS, we set $N_l=6$ for GPT-2 and Qwen2.5-0.5B-Instruct, and $N_l=8$ for LLaMA-3.2-1B-Instruct. 
More detailed configurations are provided in Appendix \ref{sec:app_b}.
\begin{table}[ht]
\caption{\textbf{Scalability to a larger backbone model.} We report answer accuracy (Acc.) and output token length (Len.) on the four aforementioned benchmarks using LLaMA-3.2-3B-Instruct. Explicit CoT method is marked with \colorbox{gray!20}{\phantom{xxx}}. The best and second best accuracy among latent reasoning approaches are highlighted in \textbf{bold} and \underline{underline}, respectively.}
\label{tab:larger_model}
\centering
\resizebox{1.0\linewidth}{!}{
\begin{tabular}{lccccccccc}
\toprule
\multicolumn{1}{c}{\multirow{3}{*}{\textbf{Method}}} & \multicolumn{2}{c}{\textbf{In-domain}} & \multicolumn{5}{c}{\textbf{Out-of-domain}}& \multicolumn{2}{c}{\multirow{2}{*}{\textbf{Average}}} \\ \cmidrule(rl){2-3} \cmidrule(rl){4-8}
 & \multicolumn{2}{c}{GSM8K-Aug} & \multicolumn{1}{c}{GSM-Hard} & \multicolumn{1}{c}{SVAMP} & \multicolumn{1}{c}{MultiArith} & \multicolumn{2}{c}{Average} &\\ 
\cmidrule(rl){2-3} \cmidrule(rl){4-4} \cmidrule(rl){5-5} \cmidrule(rl){6-6} \cmidrule(rl){7-8} \cmidrule(rl){9-10}
 & Acc. (\%) & Len. & Acc. (\%)  & Acc. (\%)& Acc. (\%)& Acc. (\%)& Len.& Acc. (\%)& Len. \\ 
\cmidrule(rl){1-10}
\rowcolor{gray!20} CoT-SFT  & 71.27 & 32.95 & 20.52 & 78.00 & 100.00 & 66.17 & 27.89 & 67.45 & 29.15 \\
Coconut & 11.98 & 12.25 & 2.35 & 14.67 & 13.33 & 10.12 & 12.23 & 10.58 & 12.24    \\
CoLaR-2 &57.39 & 12.89 & 13.87 & 69.00 & 96.67 & 59.85 & 9.88 & 59.23 & 10.63   \\
CODI & 54.28 & 13.20 & 12.96 & 67.00& \textbf{100.00} & 59.99 & 13.48 & 58.56 & 13.41 \\
SIM-CoT & \underline{61.41} & 13.19 & \underline{14.03} &\textbf{71.67} & \underline{97.22} & \underline{60.97} &13.53& \underline{61.08} & 13.45 \\
PMPS(Ours) & \textbf{63.46} & 13.17 & \textbf{14.40} & \underline{71.33} & \textbf{100.00} & \textbf{61.91} & 13.48&\textbf{62.30(+1.22)} & 13.40 \\ 
\bottomrule
\end{tabular}
\vspace{-10pt}
}
\end{table}

\begin{figure}[t]
    \begin{minipage}[t]{0.55\textwidth}
        \vspace{0pt}  
        \captionof{table}{\textbf{Scalability to more challenging task.} We conduct experiments on MATH using LLaMA-3.2-3B-Instruct and Qwen3-4B-Instruct. The best accuracy among latent approaches is highlighted in \textbf{bold}.}
        \label{tab:math}
        \centering
        \resizebox{1\linewidth}{!}{
            \begin{tabular}{l|cc|cc}
            \toprule
            \multicolumn{1}{c|}{\multirow{2}{*}{Method}} & \multicolumn{2}{c|}{Llama-3.2-3B} & \multicolumn{2}{c}{Qwen3-4B} \\
            & Acc. (\%) & Len. & Acc. (\%)& Len.\\
            \cmidrule(rl){1-5}
            \rowcolor{gray!20} COT-SFT  & 21.00& 119.69 &38.60 & 132.26  \\
            CoLaR-2   & 6.20  &  54.27  & 18.40 & 54.30  \\
            CODI   & 8.80  &  14.85  & 17.60   & 17.61  \\
            PMPS(Ours)  & \textbf{11.00}  &  14.95 & \textbf{19.60}  & 17.60 \\
            \bottomrule
            \end{tabular}
            }
    \vspace{-10pt}
    \end{minipage}
    \hfill
    \begin{minipage}[t]{0.43\textwidth}
        \vspace{0pt}  
        \centering
        \includegraphics[width=0.95\linewidth]{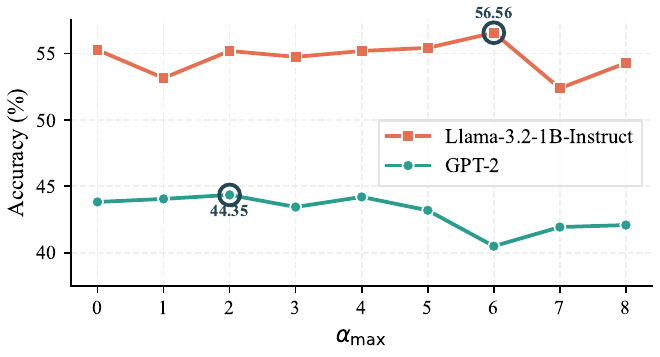}
        \caption{\textbf{Ablation Study on PSA maximum enforcement strength.} The best configurations are marked with circles. }
        \label{fig:psa}
    \vspace{-10pt}
    \end{minipage}
    \vspace{-10pt}
\end{figure}

\subsection{Main Result}
Table~\ref{tab:main} presents the main results on one in-domain benchmark (GSM8K-Aug) and three out-of-domain benchmarks (GSM-Hard, SVAMP, MultiArith), with three backbone models from different model families: GPT-2, Qwen2.5-0.5B-Instruct, and LLaMA-3.2-1B-Instruct.

\paragraph{Overall performance.}
On GPT-2, PMPS outperforms the two leading latent reasoning baselines, CODI and SIM-CoT, by 3.49\% and 2.27\% on GSM8K-Aug, and by 4.33\% and 2.86\% in average accuracy. Remarkably, PMPS even exceeds the explicit reasoning upper bound CoT-SFT by 3.96\% in average accuracy while using only 47.1\% of the output tokens, demonstrating that latent reasoning with effective process supervision can surpass explicit CoT at significantly higher inference efficiency. On Qwen2.5-0.5B-Instruct and LLaMA-3.2-1B-Instruct, PMPS consistently outperforms all latent reasoning baselines (+3.44\% and +1.23\%).
These gains across diverse model architectures and scales confirm the effectiveness of our proposed structural process supervision method. 

\paragraph{Out-of-domain (OOD) generalization.}
The OOD results show that PMPS generalizes robustly to unseen benchmarks: PMPS achieves an average accuracy of 51.82\%, 54.39\%, and 58.10\% across three backbone models, surpassing the strongest latent baseline SIM-CoT by 3.06\%, 3.04\%, and 1.46\%. The superior performance on OOD benchmarks confirms that our proposed PMPS helps internalize genuine arithmetic reasoning rather than shallow pattern matching.
 \paragraph{Inference efficiency.}
Compared to explicit CoT-SFT, PMPS reduces the output token length to under 50\% across all three models (47.1\%, 38.6\% and 47.2\%, respectively) while achieving comparable or superior accuracy.
Critically, PMPS maintains output token length comparable to CODI and SIM-CoT, verifying that the accuracy improvements stem entirely from superior training-time structural process supervision rather than increased inference computation. Moreover, this structural supervision is lightweight: the projection head and prototypes only introduce less than 2M extra parameters ($<1\%$ of the base model), whereas SIM-CoT's auxiliary decoder effectively doubles the trainable parameters during training.

\subsection{Scaling to Larger Models and Harder Task}
To further validate the scalability of PMPS, we conduct two additional groups of experiments: scaling to a larger backbone model and training on a more challenging dataset.
\paragraph{Larger model.} As shown in Table~\ref{tab:larger_model}, we scale up the backbone model to LLaMA-3.2-3B-Instruct \citep{grattafiori2024llama} and evaluate performance across the aforementioned benchmarks. PMPS achieves the highest average accuracy among all latent reasoning baselines, outperforming CODI and SIM-CoT by +3.74\% and +1.22\%, respectively, while maintaining comparable output length. Notably, PMPS demonstrates marked improvements over CODI on the in-domain GSM8K-Aug benchmark (+9.18\%) and reduces the generation token length by over 54\% relative to explicit CoT method. These results confirm that our proposed structural prototype-based alignment in PMPS remains highly effective as the model capacity scales up.


\paragraph{More challenging task.} As presented in Table~\ref{tab:math}, we conduct training and evaluation on MATH, which consists of highly challenging competition-level problems with substantially longer reasoning chains. We employ two larger backbone models: LLaMA-3.2-3B-Instruct and Qwen3-4B-Instruct-2507 \citep{yang2025qwen3}. Detailed experiment configurations are illustrated in Appendix \ref{app:math_settings}.
On LLaMA-3.2-3B-Instruct and Qwen3-4B-Instruct-2507, PMPS respectively achieves an accuracy gain of +2.20\% and +2.00\% compared to CODI, while significantly compressing the output length by 87.51\% and 86.69\% compared to explicit CoT. 
Notably, the CoT annotations in MATH are formulated as free-form solutions without explicit step delimiters, which precludes the application of methods relying on strict one-to-one latent-to-step alignment (e.g., Coconut, SIM-CoT). In contrast, PMPS operates via soft prototype matching and remains agnostic to specific formatting, enabling its direct application to any CoT data.
\begin{figure}[t]
    \centering
    \includegraphics[width=\linewidth]{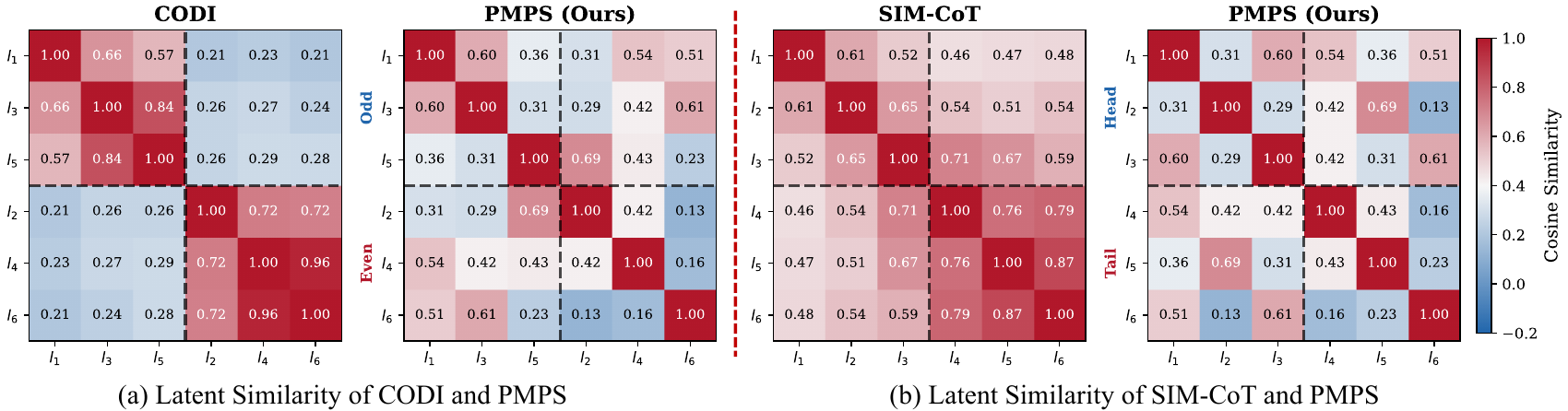}
    \caption{\textbf{Average pairwise latent cosine similarity heatmaps.} The latent embeddings of CODI and SIM-CoT exhibit pronounced odd-even and tail collapse, respectively, whereas PMPS shows neither pattern of collapse.}
    \label{fig:latent_nanlysis}
    \vspace{-10pt}
\end{figure}

\subsection{Latent Embedding Analysis}
To further investigate the reasons behind PMPS’s superior performance, we directly analyze the pairwise cosine similarity among the $N_l{=}6$ latent embeddings on randomly sampled 100 GSM8K-Aug test instances using GPT-2. The analysis reveals that CODI and SIM-CoT suffer from distinct patterns of \textbf{representation collapse}, whereas PMPS maintains well-diversified representations free from either collapse mode. Additionally, we conduct the latent embedding interpretability and comparision experiments, which are provided in Appendix \ref{sec:app_c}.

\paragraph{PMPS resolves the odd-even collapse in CODI.}

Figure~\ref{fig:latent_nanlysis} (a) presents the average pairwise cosine similarity matrices for CODI (left) and PMPS (right), where tokens are grouped into odd-indexed ($l_1, l_3, l_5$) and even-indexed ($l_2, l_4, l_6$) subsets. For CODI, two homogeneous diagonal blocks emerge: within-odd similarity averages 0.693, within-even reaches 0.800, yet cross-group similarity drops to 0.250. 
This pattern persists across all 100 samples.
This phenomenon arises because CODI's $\mathcal{L}_{dist}$ constrains only the answer-position hidden state, leaving the latent embeddings unsupervised. The autoregressive mechanism then converges to a degenerate solution where each token diverges from its predecessor, forming two alternating representations. Our PMPS avoids this collapse through two complementary mechanisms: the Sinkhorn equipartition constraint assigns different latent tokens to distinct prototypes, and the bidirectional cross-prediction loss forces each token to encode unique CoT semantics. As shown in the right panel, PMPS exhibits no block-diagonal separation between odd and even groups, confirming preserved representation diversity.

\paragraph{PMPS eliminates the tail collapse in SIM-CoT.} Figure~\ref{fig:latent_nanlysis} (b) compares the average pairwise cosine similarity matrices of SIM-CoT (left) and PMPS (right). For SIM-CoT, similarity increases from head to tail, with the tail group ($l_4, l_5, l_6$) forming a prominent high-similarity block (average 0.805) that far exceeds the head group (average 0.591). 
This positive gap is observed in 85\% of test samples.
This stems from SIM-CoT's rigid one-to-one matching, which binds each latent token to exactly one CoT step. In our training data, 99.0\% of samples have fewer steps than $N_l{=}6$, leaving the tail latent tokens without corresponding supervision and collapsing into redundant representations. PMPS resolves this through soft many-to-many matching under the equipartition constraint, enabling each latent token to flexibly aggregate CoT content regardless of position.
As shown in the right panel, PMPS displays no tail-concentrated block, with the tail group similarity even lower than head group, demonstrating that all six latent tokens encode diverse information.

\subsection{Ablation Study}
In this section, we conduct ablation studies to investigate the effects of different PMPS components, the PSA maximum enforcement strength, the number of latent embeddings and prototypes, and supervision loss weighting coefficient $\lambda_{proto}$. The number of latent embeddings and prototypes results, and $\lambda_{proto}$ results are provided in Appendix \ref{sec:app_d}.

\paragraph{Ablation on PMPS components.} 
\begin{wraptable}{r}{0.6\textwidth}
  \vspace{-10pt}  
  \centering
  \caption{\textbf{Ablation Study on PMPS components using GPT-2.} Different modules are removed to assess their contributions to the final performance.}
    \label{tab:abla}
    \resizebox{1\linewidth}{!}{
    \begin{tabular}{l|c|c|c}
    \toprule
    \multicolumn{1}{c|}{\multirow{2}{*}{Method}} & GSM8K-Aug & OOD Average & Average\\
     & Acc. (\%) & Acc. (\%) & Acc. (\%)\\
    \cmidrule(rl){1-4}
    \textbf{PMPS}  & \textbf{44.35} & \textbf{51.82} & \textbf{49.95} \\
    \;- w/o $\mathcal{L}_{l2c} $   & 39.42 & 48.42 & 46.17(-3.78) \\
    \;- w/o $\mathcal{L}_{c2l} $   & 42.23  & 49.79 & 47.90(-2.05) \\
    \;- w/o PSA   & 43.82  & 50.98  & 49.19(-0.76) \\
    \;- w/o prototype   & 41.70  & 48.04 & 46.46(-3.49) \\
    \bottomrule
    \end{tabular}
    }
  \vspace{-10pt}  
\end{wraptable}
The results in Table~\ref{tab:abla} isolate the contribution of each component. 
Removing ($L_{l2c}$) reduces average accuracy by 3.78\%, as this objective updates the shared projection head and prototypes by matching predictions from CoT representations to latent assignment codes, thereby shaping the shared prototype space that mediates process supervision. Removing ($L_{c2l}$) decreases average accuracy by 2.05\%, as this objective directly supervises latent representations by encouraging their aggregated prototype predictions to match the corresponding CoT assignment codes. Together, these results support the complementary roles of direct latent supervision through ($L_{c2l}$) and shared prototype-space learning through ($L_{l2c}$), with their combination yielding the best performance.
Removing PSA leads to a consistent decline ($-0.76\%$ on Average), validating that the curriculum-scheduled positional prior provides necessary sequential structure guidance that pure semantic matching alone cannot recover. Removing prototypes (details in Appendix \ref{app:component}) also yields a substantial drop($-3.49\%$ on Average), confirming that the categorical structure imposed by prototypes, where alignment is mediated through shared clusters, is essential for establishing meaningful many-to-many correspondences between the latent and explicit CoT embeddings.

\paragraph{Ablation on maximum enforcement strength.}Figure~\ref{fig:psa} studies the effect of $\alpha_{\max}$ on GPT-2 and Llama-3.2-1B-Instruct. Both models benefit from a moderate $\alpha_{\max}$: GPT-2 peaks at $\alpha_{\max}=2$ (44.35\%), while Llama-3.2-1B achieves its best at $\alpha_{\max}=6$ (56.56\%). Excessively strong enforcement hurts performance on both models, as overly rigid positional constraints disrupt the flexibility of semantic-driven matching. Notably, the optimal $\alpha_{\max}$ scales with model capacity: GPT-2 (124M) favors a lighter prior, while Llama-3.2-1B (1B) benefits from stronger guidance, suggesting that larger models can better leverage structural priors before cosine annealing releases the constraint. 

\section{Conclusion}
We propose PMPS, a novel training paradigm for latent reasoning that performs structural process-level supervision on latent embeddings via soft alignment. Experimental results demonstrate that across diverse model families, model sizes, and task difficulties, our method consistently achieves SOTA performance among latent reasoning approaches while reducing the output token length by more than 50\% compared to explicit CoT. These findings highlight the strong generalization, robustness, and scalability of our approach. Furthermore, we conduct an in-depth analysis from the latent representation perspective, offering insights into why PMPS consistently outperforms existing baselines. However, the current evaluation of PMPS is mainly limited to mathematical reasoning. Future work should further explore the effectiveness of PMPS on a broader range of reasoning tasks.

\bibliography{iclr2027_conference}
\bibliographystyle{iclr2027_conference}

\appendix
\section{Method Supplement} \label{app:method}
\subsection{Overall PMPS Training} \label{sec:psedo}
Algorithm~\ref{alg:pmps} summarizes the training procedure of PMPS
with Progressive Sequential Alignment (PSA). Each training example
contains a question $x$, an explicit reasoning chain $c$, and an
answer $y$. The teacher and student tasks share the same language
model $f_\theta$. The prototype projector $g_\omega$ and the
prototype matrix $P\in\mathbb{R}^{K\times d}$ are shared across
the two representation views.

\begin{algorithm}[t]
\caption{Training PMPS with Progressive Sequential Alignment}
\label{alg:pmps}
\small
\begin{algorithmic}[1]
\Require Dataset $\mathcal D$; model $f_\theta$;
    projector $g_\omega$; prototypes $P$; latent length $N_l$.
\Require Training steps $T$; temperatures $\tau,\tau_m$;
    Sinkhorn parameters $\epsilon,J$; PSA strength $\alpha_{\max}$;
    process loss weight $\lambda_{\mathrm{proto}}$.
\For{$t=0,\ldots,T-1$}
    \State Sample $\mathcal B=\{(x_b,c_b,y_b)\}_{b=1}^{B}$.
    \State Run the teacher task on $(x_b,c_b,y_b)$;
        compute $\mathcal L_{\mathrm{ce}}^{\mathrm{tea}}$
        on $(c_b,y_b)$.
    \State In a separate no-gradient teacher pass, extract CoT
        predictor states $H_c$ and answer-anchor targets
        $U^{\mathrm{tea}}$.
    \State Run $N_l$ recurrent latent steps on each $x_b$;
        collect $H_l^b=[l_1^b;\ldots;l_{N_l}^b]$.
    \State Teacher-force the student answer $y_b$;
        obtain $\mathcal L_{\mathrm{ce}}^{\mathrm{stu}}$
        and answer-anchor states $U^{\mathrm{stu}}$.
    \State Compute $\mathcal L_{\mathrm{base}}$ from both task losses
        and distillation between $U^{\mathrm{stu}}$
        and $U^{\mathrm{tea}}$.
    \State $P\gets\operatorname{NormalizeRows}(P)$.
    \For{$v\in\{l,c\}$}
        \State $Z_v\gets
            \operatorname{NormalizeRows}(g_\omega(H_v))$;
            $S_v\gets Z_vP^\top$.
        \State $Q_v^{\mathrm{valid}}\gets
            \Call{Sinkhorn}{
                \operatorname{ValidRows}(S_v),\epsilon,J}$.
        \State Restore $Q_v^{\mathrm{valid}}$
            to the per-example layout $Q_v$;
            set padded entries to zero.
        \State $\phi_v\gets
            \operatorname{softmax}_{\mathrm{row}}(S_v/\tau)$.
    \EndFor
    \State $\alpha_t\gets\frac{\alpha_{\max}}{2}
        [1+\cos(\pi t/T)]$.
    \For{$b=1,\ldots,B$}
        \State Construct the positional distance matrix $D^b$
            using Eq.~\ref{eq:psa} with strength $\alpha(t)$.
        \State $\hat M^b\gets
            Q_c^b(Q_l^b)^\top/\tau_m-\alpha_tD^b$.
        \State $A_{c2l}^b\gets
            \operatorname{softmax}_{\mathrm{row}}(\hat M^b)$;
            $A_{l2c}^b\gets
            \operatorname{softmax}_{\mathrm{row}}
            ((\hat M^b)^\top)$.
        \State $\widetilde\phi_c^b\gets A_{c2l}^b\phi_l^b$;
            $\widetilde\phi_l^b\gets A_{l2c}^b\phi_c^b$.
    \EndFor
    \State $\mathcal L_{c2l}\gets
        -\frac{1}{\sum_b N_c^b}
        \sum_{b,j,k}Q_{c,jk}^b
        \log\widetilde\phi_{c,jk}^b$.
    \State $\mathcal L_{l2c}\gets
        -\frac{1}{BN_l}
        \sum_{b,i,k}Q_{l,ik}^b
        \log\widetilde\phi_{l,ik}^b$;
        $\mathcal L_{\mathrm{proto}}\gets
        \frac{1}{2}(\mathcal L_{c2l}+\mathcal L_{l2c})$.
    \State $\mathcal L_{\mathrm{total}}\gets
        \mathcal L_{\mathrm{base}}
        +\lambda_{\mathrm{proto}}\mathcal L_{\mathrm{proto}}$.
    \State Update $\theta,\omega,P$ using
        $\nabla_{\theta,\omega,P}\mathcal L_{\mathrm{total}}$.
\EndFor
\State \Return Trained student model.
\end{algorithmic}
\end{algorithm}
All matching operations and loss sums in Algorithm~\ref{alg:pmps}
are restricted to valid positions. In the padded implementation,
invalid matching destinations are masked before softmax, and
padded source positions are excluded from loss reduction.
Each directional loss is averaged over its valid target tokens
across the minibatch, rather than averaging the per-example losses.
Mixture probabilities are clamped below at $10^{-12}$ before
taking logarithms.

\paragraph{Gradient flow and inference.}
Both the teacher targets and the Sinkhorn codes are detached,
so the matching weights are also treated as fixed targets during
each update.
Gradients from $\mathcal L_{\mathrm{proto}}$ flow through the
prototype predictions to $g_\omega$ and $P$, and through the
student representations to the language model.
They do not flow through $H_c$ to the teacher computation,
while the shared language model still receives gradients from
the teacher cross-entropy.
At inference, only the student computation is retained.
The prototype projector $g_\omega$ and $P$ are discarded;
the latent recurrence projector remains part of the student model.

\subsection{Sinkhorn-Knopp Details} \label{sec:sink}
Algorithm~\ref{alg:pmps-sinkhorn} computes the detached prototype
codes used in Algorithm~\ref{alg:pmps}.
Its input $S\in\mathbb{R}^{M\times K}$ contains the scores of all
valid tokens from one representation view, where $M$ is the valid
token count and $K$ is the number of prototypes.
The procedure is applied independently to the latent and
explicit CoT views.

\begin{algorithm}[t]
\caption{Sinkhorn--Knopp Assignment over Valid Minibatch Tokens}
\label{alg:pmps-sinkhorn}
\small
\begin{algorithmic}[1]
\Require Scores $S\in\mathbb{R}^{M\times K}$ with $M>0$;
    regularization $\epsilon$; iterations $J$.
\Ensure Detached assignments $Q\in\mathbb{R}^{M\times K}$.
\Function{Sinkhorn}{$S,\epsilon,J$}
    \State Disable gradient tracking for the following operations.
    \State $\delta\gets 10^{-8}$;
        $R\gets\exp(S/\epsilon)^\top$
        \Comment{$R\in\mathbb{R}^{K\times M}$}
    \State $R\gets R/(\sum_{k,m}R_{km}+\delta)$.
    \For{$r=1,\ldots,J$}
        \State $a_k\gets\sum_{m=1}^{M}R_{km}$ for all $k$.
        \State $R_{km}\gets R_{km}/[K(a_k+\delta)]$
            for all $k,m$.
        \State $b_m\gets\sum_{k=1}^{K}R_{km}$ for all $m$.
        \State $R_{km}\gets R_{km}/[M(b_m+\delta)]$
            for all $k,m$.
    \EndFor
    \State \Return $Q\gets MR^\top$.
\EndFunction
\end{algorithmic}
\end{algorithm}

\paragraph{Balanced assignment.}
Exponentiating the scores with regularization parameter $\epsilon$
produces a positive assignment matrix.
In the transposed matrix $R$, rows index prototypes and columns
index tokens.
The two normalization steps alternately impose prototype mass
$1/K$ and token mass $1/M$.
After transposing and rescaling by $M$, we obtain the assignment matrix $Q\in\mathbb{R}^{M\times K}$.
\begin{equation}
    Q\mathbf{1}_{K}=\mathbf{1}_{M},
    \qquad
    Q^\top\mathbf{1}_{M}
        =\frac{M}{K}\mathbf{1}_{K},
    \label{eq:app-sinkhorn-marginals}
\end{equation}
Here, $\mathbf{1}_{n}$ denotes the all-ones vector of length $n$.
The first constraint normalizes each token's assignment
distribution, while the second enforces uniform aggregate
mass across prototypes.
In practice, these constraints are approximately satisfied
due to the finite number of iterations and numerical stabilization.

\paragraph{Implementation details.}
We use $J=3$ iterations and $\epsilon=0.05$.
With a finite number of iterations, token-wise normalization is
approximately satisfied after the final token-normalization step,
whereas uniform prototype usage is approximate rather than exact.
Padding tokens are removed before assignment and restored as
zero rows afterward.
In distributed training, the current implementation applies
Sinkhorn within each device's minibatch, without gathering scores
across devices.
No gradients are propagated through this procedure; the returned
codes supervise the differentiable prototype predictions.

\section{Benchmarks and Baselines} \label{sec:app_a}
\subsection{Benchmarks}\label{sec:bench}
\paragraph{GSM8K-Aug.} GSM8K-Aug expands the original GSM8K training set from approximately 7.5K to 385K examples through GPT-4-based data generation, while retaining the standard 1,319-example test split. Unlike the natural language reasoning chains in GSM8K, the chain-of-thought annotations in GSM8K-Aug are represented as structured sequences of arithmetic expressions (e.g., \texttt{<<12*3=36>><<9*2=18>><<36+18=54>>}), where each expression is logically derived from preceding steps. This format facilitates training latent reasoning models by providing concise yet faithful step-by-step supervision.

\paragraph{GSM-Hard.} GSM-Hard is a more challenging variant of GSM8K, created by replacing the numerical values in the original problems with larger numbers that require more complex arithmetic operations. This dataset is designed to evaluate the arithmetic robustness and compositional reasoning capabilities of language models under increased computational difficulty.

\paragraph{SVAMP.} SVAMP (Simple Variations on Arithmetic Math word Problems) is a challenge set of elementary-level math word problems that tests sensitivity to structural variations in problem statements. By systematically varying question types, equation structures, and entity associations, SVAMP exposes models that rely on shallow heuristics rather than genuine mathematical reasoning.

\paragraph{MultiArith.} MultiArith is a benchmark consisting of multi-step arithmetic word problems that require two or more mathematical operations to solve. It evaluates a model's ability to correctly decompose complex problem narratives into sequential reasoning steps involving basic arithmetic operations such as addition, subtraction, multiplication, and division.

\paragraph{MATH.} MATH is a comprehensive benchmark comprising 12,500 competition-level mathematics problems spanning seven subjects including algebra, geometry, number theory, and combinatorics. Problems are annotated with difficulty levels from 1 to 5, and each solution requires multi-step reasoning with formal mathematical derivations, making it a rigorous testbed for advanced mathematical reasoning capabilities.

\subsection{Baselines}\label{sec:baseline}
\paragraph{CoT-SFT.} CoT-SFT is a standard baseline that directly fine-tunes the language model on (Question, Chain-of-Thought, Answer) triples via supervised fine-tuning, where the model learns to generate explicit step-by-step reasoning tokens before producing the final answer. It serves as the upper-bound reference for explicit reasoning performance, against which latent reasoning methods are compared.

\paragraph{Coconut.} Coconut proposes a multi-stage curriculum learning strategy to train latent reasoning: starting from standard CoT data, it progressively replaces explicit language reasoning steps with continuous thoughts (the last hidden states of the LLM fed back directly as input embeddings), stage by stage, until all reasoning steps are internalized into the continuous latent space. This curriculum-based approach allows the model to gradually transition from explicit language reasoning to fully latent reasoning, avoiding the difficulty of learning continuous thought representations from scratch.

\paragraph{CoLaR-2.} CoLaR-2 dynamically compresses explicit reasoning chains into the latent space through a two-stage training approach: first, supervised fine-tuning with an auxiliary next compressed embedding prediction objective that merges consecutive token embeddings using a variable compression factor; second, reinforcement learning that leverages the latent head's non-deterministic nature to explore diverse reasoning paths. In our experiments, we set the compression rate to 2. This framework allows the model to adjust reasoning speed at inference time by simply prompting the desired compression factor, achieving significant reductions in reasoning chain length with minimal performance degradation. 

\paragraph{Latent-SFT.} Latent-SFT frames latent reasoning as a "chain of superposition", where each latent token carries higher entropy and encodes a superposition of multiple reasoning trajectories rather than merely compressing a single reasoning path. It introduces a unified framework addressing challenges at three levels: constraining hidden states within the pre-trained vocabulary space (Latent-Vocab), constructing semantically compact and sufficient latent chains via Induction-Supervision Masking (Latent-Chain), and employing stochastic Gumbel-Softmax optimization to guide the model toward generalizable solutions (Latent-Optim).

\paragraph{CODI.} CODI compresses explicit Chain-of-Thought reasoning into a continuous latent space through a self-distillation framework, jointly training a teacher task (explicit CoT) and a student task (implicit CoT) while aligning their hidden states at a designated token position. 

\paragraph{Sim-CoT} Sim-CoT addresses the latent instability problem in implicit CoT methods where latent representations become homogeneous and lose semantic diversity as the number of reasoning tokens increases by introducing step-level supervision via an auxiliary decoder during training that aligns each implicit token with its corresponding explicit reasoning step. The auxiliary decoder is removed at inference time, preserving the token efficiency of implicit CoT with no added overhead, while also providing interpretability through per-step projection onto the explicit reasoning vocabulary.

\section{Implementation Details}\label{sec:app_b}
In this section, we present the detailed hyperparameter configurations used in GSM8K-Aug experiments, the MATH experiments in Table \ref{tab:math}, and the component ablations in Table \ref{tab:abla}. Across all experiments, we adopt the AdamW optimizer with a weight decay of 0.1 and a warmup ratio of 0.03. For CoT-SFT, CoLaR-2, CODI, SIM-CoT, and our proposed PMPS, we apply LoRA-based fine-tuning with the LoRA rank set to 128 and the LoRA alpha set to 32. All models are trained on 8 NVIDIA A100 GPUs, and all evaluations are conducted on a single NVIDIA A100 GPU.
\subsection{Experimental Settings on GSM8K-Aug}
\paragraph{CoT-SFT.} For GPT-2, the batch size is set to 256 and the learning rate to $8\times10^{-4}$, trained for 40 epochs. For Qwen2.5-0.5B-Instruct and Llama-3.2-1B-Instruct, the batch size is set to 512 with learning rates of $8\times10^{-4}$ and $9\times10^{-4}$, respectively, trained for 20 epochs.

\paragraph{Coconut.} All models use a batch size of 128 and a learning rate of $1\times10^{-4}$. Each stage replaces one CoT step with two latent embeddings, with a maximum of 5 stages, trained for 20 epochs in total. All other parameters remain consistent with the original paper.

\paragraph{CoLaR-2.} All models are configured with a compression rate of 2, a batch size of 256, and a learning rate of $1\times10^{-4}$, trained for 50 epochs. All other parameters follow the original paper.

\paragraph{Latent-SFT.} In Stage 2, the Gumbel temperature and noise scale are set to 1.0 for all models. All models are configured with a batch size of 512 and a learning rate of $3\times10^{-4}$, trained for 50 epochs. All other stages and parameters remain consistent with the original paper.

\paragraph{CODI and Sim-CoT.} For GPT-2, Qwen2.5-0.5B-Instruct, and Llama-3.2-1B-Instruct, the learning rates are set to $3\times10^{-3}$, $8\times10^{-4}$, and $9\times10^{-4}$, respectively, with corresponding distillation loss factors of 1.0, 1.0, and 20.0. All other parameters follow the original paper.

\paragraph{PMPS.} For training on GSM8K-Aug, the regularization parameter $\epsilon$ is set to 0.05, the matching temperature $\tau_m$ to 0.3, the prediction temperature $\tau$ to 0.1, and the prototype dimension $d$ to 128. To improve prototype utilization and stability, Sinkhorn assignment is performed at the batch level rather than on individual samples, with the number of iterations set to 3. For GPT-2 and Qwen2.5-0.5B-Instruct, $\lambda_{proto}$ is set to 0.5, the batch size to 256, the numbers of latents and prototypes to 6 and 24 respectively, and the maximum enforcement strength to 2. For Llama-3.2-1B-Instruct, $\lambda_{proto}$ is set to 0.3, the batch size to 512, the number of latents and prototypes to 8 and 32, respectively, and the maximum enforcement strength to 6. All remaining parameters are consistent with CODI.
\subsection{Experimental Settings on MATH}
\label{app:math_settings}

For experiments on MATH in Table \ref{tab:math}, we use Llama-3.2-3B-Instruct and Qwen3-4B-Instruct-2507 as the backbone models. 
We retain the complete solution text as the explicit CoT supervision without segmenting it into predefined reasoning steps. 

\paragraph{CoT-SFT.} Both Llama-3.2-3B-Instruct and Qwen3-4B-Instruct-2507 models are trained with learning rate of $8\times10^{-4}$ and batch size of 256 for 50 epochs.

\paragraph{CoLaR-2.} Both models are trained with learning rate of $1\times10^{-4}$ for 100 epochs. Other configurations are the same with GSM8K-Aug experiments.

\paragraph{PMPS and CODI.} Both methods use $N_l=8$ latent embeddings and a latent recurrence projector with hidden dimension 3072 and trained for 100 epochs.
For Llama-3.2-3B-Instruct, we use a learning rate of $9\times10^{-4}$ and a batch size of 256.
For Qwen3-4B-Instruct-2507, we use a learning rate of $5\times10^{-4}$ and a batch size of 256.
For PMPS, the $\lambda_{proto}$ is set to 0.3 and the number of prototypes is set to 32.

\begin{figure}[t]
    \centering
    \includegraphics[width=\linewidth]{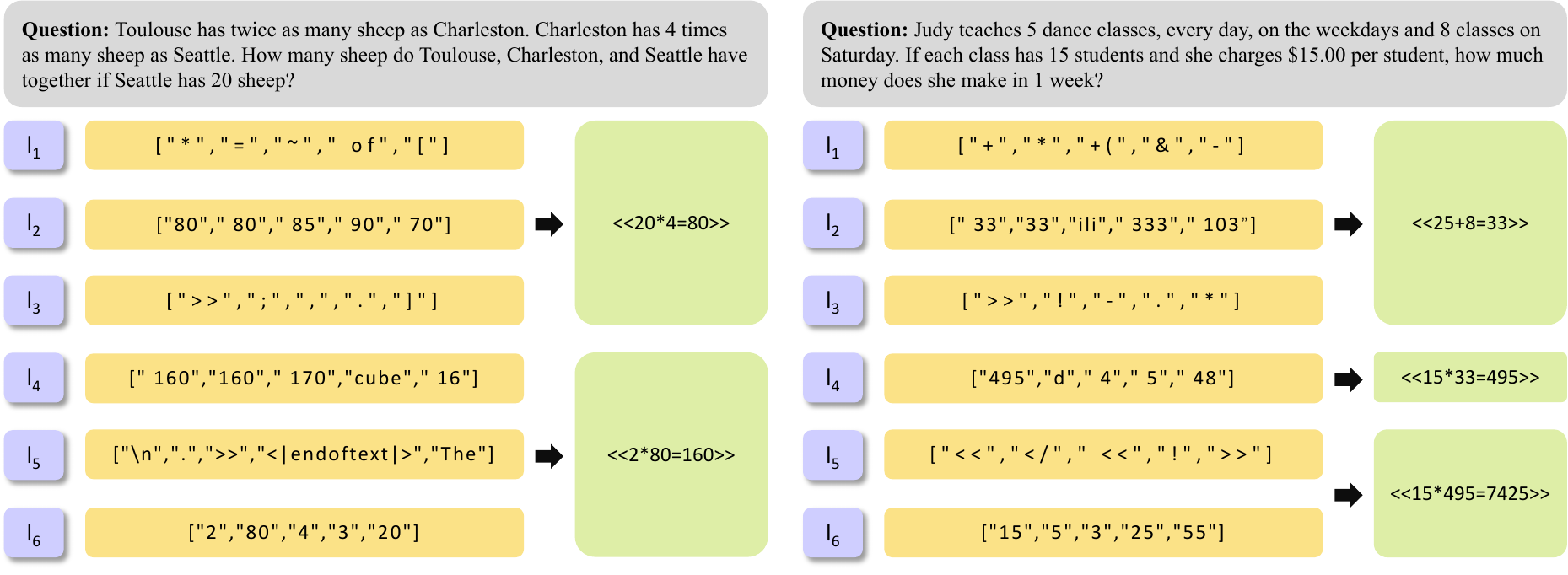}
    \caption{\textbf{Interpretability cases of PMPS.} We decode the latent embeddings and visualize the top-5 tokens with the highest probabilities, aligning them with the corresponding CoT steps. }
    \label{fig:case}
\end{figure}
\begin{figure}[t]
    \centering
    \includegraphics[width=\linewidth]{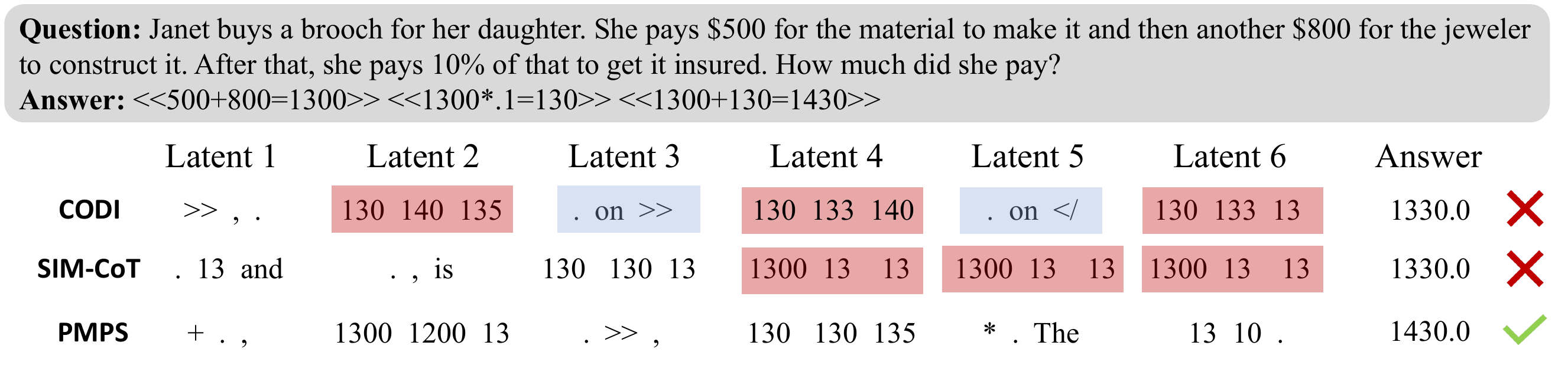}
    \caption{\textbf{Case comparision of CODI, SIM-CoT and PMPS.} All methods use GPT-2 with six latent steps. Each column lists the top-3 tokens obtained by projecting the corresponding latent state through the language model head. Colored backgrounds highlight recurring readouts across CODI’s odd/even positions and SIM-CoT’s final three positions.}
    \label{fig:case_compara}
\end{figure}

\subsection{Experimental Settings for Component Ablations} \label{app:component}
We conduct the component ablations in Table \ref{tab:abla} using GPT-2 trained on
GSM8K-Aug and evaluate on GSM8K-Aug, GSM-Hard, SVAMP, and MultiArith.
\paragraph{Removing a directional loss.}
For \text{w/o $\mathcal{L}_{l2c}$}, we remove the latent-to-CoT
cross-prediction term and retain only $\mathcal{L}_{c2l}$ for process
supervision. Conversely, \text{w/o $\mathcal{L}_{c2l}$} removes
the CoT-to-latent term and retains only $\mathcal{L}_{l2c}$.
Both variants retain the shared projection head, prototypes,
Sinkhorn assignments, and PSA. All other settings follow the
reference configuration.

\paragraph{Removing PSA.}
For \text{w/o PSA}, we set the positional enforcement strength
to zero throughout training, so the matching weights depend only
on the prototype assignment similarities. Both directional losses
and all other settings remain unchanged.

\paragraph{Removing prototypes.}
For \text{w/o prototype}, we remove the learnable prototypes and
Sinkhorn assignment and replace prototype-mediated supervision with
direct matching in the projected embedding space. We retain a shared
two-layer MLP with hidden dimension 512 and output dimension 128,
followed by $\ell_2$ normalization. Pairwise cosine similarities
between the projected CoT and latent representations, scaled by
$\tau_m=0.3$ and augmented with the positional bias, are
normalized in each direction to obtain soft many-to-many matching
weights. We replace the prototype cross-entropy objective with the mean of the two directional cosine reconstruction losses, each defined as one minus the cosine similarity
between the target and its reconstruction and averaged over valid
target positions in the minibatch. 
This variant preserves bidirectional soft matching while removing the prototype-based representation and balanced assignment mechanism.

\section{Interpretability Experiment}\label{sec:app_c}
To qualitatively examine whether the latent reasoning embeddings encode semantically meaningful information, we conduct a qualitative vocabulary-readout analysis on GSM8K test examples using GPT-2 with ($N_l=6$) latent steps. For each method, we project each latent state through its language model head and rank vocabulary tokens by their resulting probabilities. We report the top-5 tokens for the PMPS examples in Figure \ref{fig:case} and the top-3 tokens for the comparison with CODI and SIM-CoT in Figure \ref{fig:case_compara}. PMPS receives supervision through teacher CoT representations, without direct token-level reconstruction targets for its intermediate latent states. Following CODI, the final answer-revealing CoT step is excluded from the teacher reasoning chain used for distillation to avoid exposing the answer verbatim before the distillation anchor.
\subsection{Semantic Content of PMPS Latent States}
Figure \ref{fig:case} shows that the PMPS readouts contain numbers and symbols associated with the reference reasoning. In the sheep example, ($l_2$) and ($l_4$) assign high probability to (80) and (160), respectively, matching the intermediate quantities in the reference CoT. In the dance-class example, the corresponding positions expose (33) and (495). These examples illustrate that task-relevant intermediate values are accessible from the latent states through vocabulary projection.
\subsection{Comparison with Baselines}
Figure \ref{fig:case_compara} compares the three methods on the same example, whose reference solution computes ($500+800=1300$), ($1300\times0.1=130$), and ($1300+130=1430$). CODI exhibits recurring readouts within its odd- and even-indexed positions: the even positions repeatedly favor (130), while ($l_3$) and ($l_5$) share punctuation and “on”. SIM-CoT displays identical rendered top-3 token lists at ($l_4$), ($l_5$), and ($l_6$), dominated by (1300) and (13). These patterns are qualitatively consistent with the odd–even and tail similarity patterns, respectively, reported in Section 4.4. In contrast, PMPS exposes (1300) at ($l_2$) and (130) at ($l_4$), together with arithmetic operators such as “+” and “*” at other positions, and produces the correct answer (1430). Both baselines output (1330).

These examples complement the aggregate similarity analysis in Figure \ref{fig:latent_nanlysis} by illustrating task-relevant content and repeated readout patterns. Vocabulary projections provide a partial view of the latent states; they do not establish a one-to-one mapping to explicit CoT steps or a causal explanation of the final predictions.
\begin{figure}[t]
    \centering
    \includegraphics[width=1.0\linewidth]{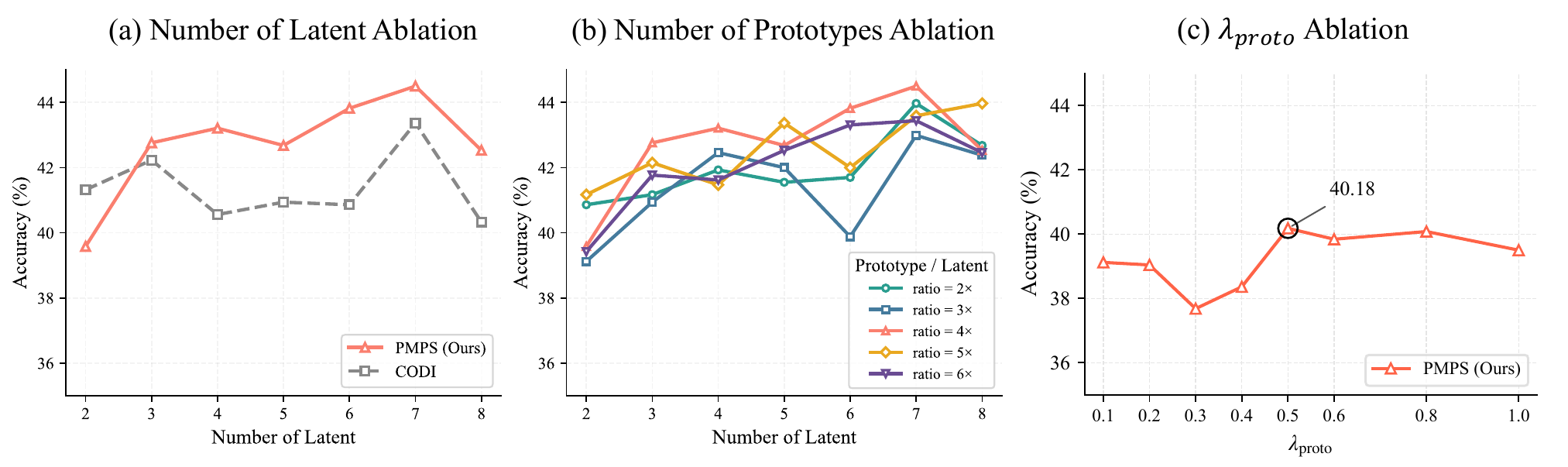}
    \caption{\textbf{Ablation studies of PMPS.} (a) Number of latent tokens, with CODI as the baseline. (b) Prototype-to-latent ratio across different latent counts. (c) Process-supervision weight ($\lambda_{proto}$) Ablation. All results are conducted with GPT-2 on GSM8K-Aug. In (c), all other hyperparameters are fixed at non-optimal settings; the best accuracy in this sweep therefore differs from the Table \ref{tab:main} result obtained with the final configuration.}
    \label{fig:abl}
\end{figure}
\section{Ablation Study}\label{sec:app_d}
\paragraph{Ablation on number of latent tokens.} Figure~\ref{fig:abl}(a) compares PMPS (with a prototype-to-latent ratio of 4$\times$) against the CODI baseline across varying numbers of latent tokens. PMPS outperforms CODI at all latent counts except for 2, where the limited number of latent tokens constrains the capacity of prototype-based alignment, leaving insufficient room for soft matching to take effect. The best performance is achieved at 6 and 7 latent tokens. Notably, CODI suffers a sharp accuracy drop at 8 latent tokens, falling to the lowest among all configurations, indicating that its answer-level distillation alone is insufficient to supervise a larger set of latent representations. PMPS, however, still maintains competitive accuracy at 8 latent tokens, demonstrating that the prototype-based alignment provides more stable supervision and enables the method to scale up effectively.

\paragraph{Ablation on number of prototypes.} Figure~\ref{fig:abl}(b) examines the impact of the prototype-to-latent ratio by varying it across \{2$\times$, 3$\times$, 4$\times$, 5$\times$, 6$\times$\}. Among all ratios, the 4$\times$ setting consistently yields the best overall performance across different latent counts. A ratio that is too small (e.g., 2$\times$ or 3$\times$) limits the expressiveness of the prototype space, providing insufficient semantic granularity for diverse reasoning patterns. Conversely, an excessively large ratio (e.g., 6$\times$) introduces redundant prototypes that may dilute the clustering signal. Based on these results, we adopt a prototype-to-latent ratio of 4$\times$ as the default setting in all other experiments.

\paragraph{Ablation on $\lambda_{proto}$.}
Figure \ref{fig:abl}(c) varies $\lambda_{proto}$ from $0.1$ to $1.0$ while keeping all other hyperparameters fixed.
Accuracy varies non-monotonically, reaching its highest observed value of $40.18\%$ at $\lambda_{proto}=0.5$.
Increasing the weight therefore does not consistently improve accuracy.
Importantly, the other hyperparameters in this sweep are not set to their optimal values, so its best result differs from the main-table result and should be interpreted as sensitivity under this fixed configuration.
\end{document}

%% file: math_commands.tex
\usepackage{amsmath,amsfonts,bm}

\def\eqref#1{equation~\ref{#1}}

\def\1{\bm{1}}

\DeclareMathAlphabet{\mathsfit}{\encodingdefault}{\sfdefault}{m}{sl}
\SetMathAlphabet{\mathsfit}{bold}{\encodingdefault}{\sfdefault}{bx}{n}



%% file: iclr2027_conference.bib
@article{achiam2023gpt,
  title={Gpt-4 technical report},
  author={Achiam, Josh and Adler, Steven and Agarwal, Sandhini and Ahmad, Lama and Akkaya, Ilge and Aleman, Florencia Leoni and Almeida, Diogo and Altenschmidt, Janko and Altman, Sam and Anadkat, Shyamal and others},
  journal={arXiv preprint arXiv:2303.08774},
  year={2023}
}

@article{grattafiori2024llama,
  title={The llama 3 herd of models},
  author={Grattafiori, Aaron and Dubey, Abhimanyu and Jauhri, Abhinav and Pandey, Abhinav and Kadian, Abhishek and Al-Dahle, Ahmad and Letman, Aiesha and Mathur, Akhil and Schelten, Alan and Vaughan, Alex and others},
  journal={arXiv preprint arXiv:2407.21783},
  year={2024}
}

@article{comanici2025gemini,
  title={Gemini 2.5: Pushing the frontier with advanced reasoning, multimodality, long context, and next generation agentic capabilities},
  author={Comanici, Gheorghe and Bieber, Eric and Schaekermann, Mike and Pasupat, Ice and Sachdeva, Noveen and Dhillon, Inderjit and Blistein, Marcel and Ram, Ori and Zhang, Dan and Rosen, Evan and others},
  journal={arXiv preprint arXiv:2507.06261},
  year={2025}
}

@article{yang2025qwen3,
  title={Qwen3 technical report},
  author={Yang, An and Li, Anfeng and Yang, Baosong and Zhang, Beichen and Hui, Binyuan and Zheng, Bo and Yu, Bowen and Gao, Chang and Huang, Chengen and Lv, Chenxu and others},
  journal={arXiv preprint arXiv:2505.09388},
  year={2025}
}

@article{liu2025deepseek,
  title={Deepseek-v3. 2: Pushing the frontier of open large language models},
  author={Liu, Aixin and Mei, Aoxue and Lin, Bangcai and Xue, Bing and Wang, Bingxuan and Xu, Bingzheng and Wu, Bochao and Zhang, Bowei and Lin, Chaofan and Dong, Chen and others},
  journal={arXiv preprint arXiv:2512.02556},
  year={2025}
}

@article{zeng2026glm,
  title={Glm-5: from vibe coding to agentic engineering},
  author={Zeng, Aohan and Lv, Xin and Hou, Zhenyu and Du, Zhengxiao and Zheng, Qinkai and Chen, Bin and Yin, Da and Ge, Chendi and Huang, Chenghua and Xie, Chengxing and others},
  journal={arXiv preprint arXiv:2602.15763},
  year={2026}
}

@article{wei2022chain,
  title={Chain-of-thought prompting elicits reasoning in large language models},
  author={Wei, Jason and Wang, Xuezhi and Schuurmans, Dale and Bosma, Maarten and Xia, Fei and Chi, Ed and Le, Quoc V and Zhou, Denny and others},
  journal={Advances in neural information processing systems},
  volume={35},
  pages={24824--24837},
  year={2022}
}

@inproceedings{shen2025codi,
  title={Codi: Compressing chain-of-thought into continuous space via self-distillation},
  author={Shen, Zhenyi and Yan, Hanqi and Zhang, Linhai and Hu, Zhanghao and Du, Yali and He, Yulan},
  booktitle={Proceedings of the 2025 Conference on Empirical Methods in Natural Language Processing},
  pages={677--693},
  year={2025}
}

@article{hao2024coconut,
  title={Training large language models to reason in a continuous latent space},
  author={Hao, Shibo and Sukhbaatar, Sainbayar and Su, DiJia and Li, Xian and Hu, Zhiting and Weston, Jason and Tian, Yuandong},
  journal={arXiv preprint arXiv:2412.06769},
  year={2024}
}

@article{wei2025sim,
  title={SIM-CoT: Supervised Implicit Chain-of-Thought},
  author={Wei, Xilin and Liu, Xiaoran and Zang, Yuhang and Dong, Xiaoyi and Cao, Yuhang and Wang, Jiaqi and Qiu, Xipeng and Lin, Dahua},
  journal={arXiv preprint arXiv:2509.20317},
  year={2025}
}

@article{cuturi2013sinkhorn,
  title={Sinkhorn distances: Lightspeed computation of optimal transport},
  author={Cuturi, Marco},
  journal={Advances in neural information processing systems},
  volume={26},
  year={2013}
}

@article{chen2025reasoning,
  title={Reasoning beyond language: A comprehensive survey on latent chain-of-thought reasoning},
  author={Chen, Xinghao and Zhao, Anhao and Xia, Heming and Lu, Xuan and Wang, Hanlin and Chen, Yanjun and Zhang, Wei and Wang, Jian and Li, Wenjie and Shen, Xiaoyu},
  journal={arXiv preprint arXiv:2505.16782},
  year={2025}
}

@article{li2025implicit,
  title={Implicit reasoning in large language models: A comprehensive survey},
  author={Li, Jindong and Fu, Yali and Fan, Li and Liu, Jiahong and Shu, Yao and Qin, Chengwei and Yang, Menglin and King, Irwin and Ying, Rex},
  journal={arXiv preprint arXiv:2509.02350},
  year={2025}
}

@misc{zhu2025surveylatentreasoning,
      title={A Survey on Latent Reasoning}, 
      author={Rui-Jie Zhu and Tianhao Peng and Tianhao Cheng and Xingwei Qu and Jinfa Huang and Dawei Zhu and Hao Wang and Kaiwen Xue and Xuanliang Zhang and Yong Shan and Tianle Cai and Taylor Kergan and Assel Kembay and Andrew Smith and Chenghua Lin and Binh Nguyen and Yuqi Pan and Yuhong Chou and Zefan Cai and Zhenhe Wu and Yongchi Zhao and Tianyu Liu and Jian Yang and Wangchunshu Zhou and Chujie Zheng and Chongxuan Li and Yuyin Zhou and Zhoujun Li and Zhaoxiang Zhang and Jiaheng Liu and Ge Zhang and Wenhao Huang and Jason Eshraghian},
      year={2025},
      eprint={2507.06203},
      archivePrefix={arXiv},
      primaryClass={cs.CL},
      url={https://arxiv.org/abs/2507.06203}, 
}

@article{zhang2026soft,
  title={Soft thinking: Unlocking the reasoning potential of llms in continuous concept space},
  author={Zhang, Zhen and He, Xuehai and Yan, Weixiang and Shen, Ao and Zhao, Chenyang and Wang, Xin},
  journal={Advances in Neural Information Processing Systems},
  volume={38},
  pages={168990--169012},
  year={2025}
}

@misc{deng2024explicit,
      title={From Explicit CoT to Implicit CoT: Learning to Internalize CoT Step by Step}, 
      author={Yuntian Deng and Yejin Choi and Stuart Shieber},
      year={2024},
      eprint={2405.14838},
      archivePrefix={arXiv},
      primaryClass={cs.CL},
      url={https://arxiv.org/abs/2405.14838}, 
}

@article{li2025latent,
  title={Latent visual reasoning},
  author={Li, Bangzheng and Sun, Ximeng and Liu, Jiang and Wang, Ze and Wu, Jialian and Yu, Xiaodong and Chen, Hao and Barsoum, Emad and Chen, Muhao and Liu, Zicheng},
  journal={arXiv preprint arXiv:2509.24251},
  year={2025}
}

@article{tan2026think,
  title={Think silently, think fast: Dynamic latent compression of llm reasoning chains},
  author={Tan, Wenhui and Li, Jiaze and Ju, Jianzhong and Luo, Zhenbo and Song, Ruihua and Luan, Jian},
  journal={Advances in Neural Information Processing Systems},
  volume={38},
  pages={4646--4668},
  year={2025}
}

@misc{deng2026llmlatentreasoningchain,
      title={LLM Latent Reasoning as Chain of Superposition}, 
      author={Jingcheng Deng and Liang Pang and Zihao Wei and Shicheng Xu and Zenghao Duan and Kun Xu and Yang Song and Huawei Shen and Xueqi Cheng},
      year={2026},
      eprint={2510.15522},
      archivePrefix={arXiv},
      primaryClass={cs.CL},
      url={https://arxiv.org/abs/2510.15522}, 
}

@inproceedings{lightman2024let,
  title={Let's verify step by step},
  author={Lightman, Hunter and Kosaraju, Vineet and Burda, Yuri and Edwards, Harrison and Baker, Bowen and Lee, Teddy and Leike, Jan and Schulman, John and Sutskever, Ilya and Cobbe, Karl},
  booktitle={International Conference on Learning Representations},
  volume={2024},
  pages={39578--39601},
  year={2024}
}

@inproceedings{wang2024math,
  title={Math-shepherd: Verify and reinforce llms step-by-step without human annotations},
  author={Wang, Peiyi and Li, Lei and Shao, Zhihong and Xu, Runxin and Dai, Damai and Li, Yifei and Chen, Deli and Wu, Yu and Sui, Zhifang},
  booktitle={Proceedings of the 62nd Annual Meeting of the Association for Computational Linguistics (Volume 1: Long Papers)},
  pages={9426--9439},
  year={2024}
}

@article{caron2020unsupervised,
  title={Unsupervised learning of visual features by contrasting cluster assignments},
  author={Caron, Mathilde and Misra, Ishan and Mairal, Julien and Goyal, Priya and Bojanowski, Piotr and Joulin, Armand},
  journal={Advances in neural information processing systems},
  volume={33},
  pages={9912--9924},
  year={2020}
}

@article{li2020prototypical,
  title={Prototypical contrastive learning of unsupervised representations},
  author={Li, Junnan and Zhou, Pan and Xiong, Caiming and Hoi, Steven CH},
  journal={arXiv preprint arXiv:2005.04966},
  year={2020}
}

@article{cobbe2021gsm8k,
  title={Training verifiers to solve math word problems},
  author={Cobbe, Karl and Kosaraju, Vineet and Bavarian, Mohammad and Chen, Mark and Jun, Heewoo and Kaiser, Lukasz and Plappert, Matthias and Tworek, Jerry and Hilton, Jacob and Nakano, Reiichiro and others},
  journal={arXiv preprint arXiv:2110.14168},
  year={2021}
}

@article{deng2023gsm8k-aug,
  title={Implicit chain of thought reasoning via knowledge distillation},
  author={Deng, Yuntian and Prasad, Kiran and Fernandez, Roland and Smolensky, Paul and Chaudhary, Vishrav and Shieber, Stuart},
  journal={arXiv preprint arXiv:2311.01460},
  year={2023}
}

@inproceedings{gao2023gsm-hard,
  title={Pal: Program-aided language models},
  author={Gao, Luyu and Madaan, Aman and Zhou, Shuyan and Alon, Uri and Liu, Pengfei and Yang, Yiming and Callan, Jamie and Neubig, Graham},
  booktitle={International conference on machine learning},
  pages={10764--10799},
  year={2023},
  organization={PMLR}
}

@inproceedings{patel2021svamp,
  title={Are NLP models really able to solve simple math word problems?},
  author={Patel, Arkil and Bhattamishra, Satwik and Goyal, Navin},
  booktitle={Proceedings of the 2021 conference of the North American chapter of the association for computational linguistics: human language technologies},
  pages={2080--2094},
  year={2021}
}

@inproceedings{roy2015multiarith,
  title={Solving general arithmetic word problems},
  author={Roy, Subhro and Roth, Dan},
  booktitle={Proceedings of the 2015 conference on empirical methods in natural language processing},
  pages={1743--1752},
  year={2015}
}

@article{radford2019gpt2,
  title={Language models are unsupervised multitask learners},
  author={Radford, Alec and Wu, Jeffrey and Child, Rewon and Luan, David and Amodei, Dario and Sutskever, Ilya and others},
  journal={OpenAI blog},
  volume={1},
  number={8},
  pages={9},
  year={2019}
}

@article{yang2024qwen2.5,
  title={Qwen2.5 technical report},
  author={Yang, An and Yang, Baosong and Zhang, Beichen and Hui, Binyuan and Zheng, Bo and Yu, Bowen and Li, Chengyuan and Liu, Dayiheng and Huang, Fei and Wei, Haoran and others},
  journal={arXiv preprint arXiv:2412.15115},
  year={2024}
}

@article{hu2022lora,
  title={Lora: Low-rank adaptation of large language models.},
  author={Hu, Edward J and Shen, Yelong and Wallis, Phillip and Allen-Zhu, Zeyuan and Li, Yuanzhi and Wang, Shean and Wang, Liang and Chen, Weizhu and others},
  journal={Iclr},
  volume={1},
  number={2},
  pages={3},
  year={2022}
}

@inproceedings{sun2019patient,
  title={Patient knowledge distillation for bert model compression},
  author={Sun, Siqi and Cheng, Yu and Gan, Zhe and Liu, Jingjing},
  booktitle={Proceedings of the 2019 conference on empirical methods in natural language processing and the 9th international joint conference on natural language processing (EMNLP-IJCNLP)},
  pages={4323--4332},
  year={2019}
}

@article{wang2020minilm,
  title={Minilm: Deep self-attention distillation for task-agnostic compression of pre-trained transformers},
  author={Wang, Wenhui and Wei, Furu and Dong, Li and Bao, Hangbo and Yang, Nan and Zhou, Ming},
  journal={Advances in neural information processing systems},
  volume={33},
  pages={5776--5788},
  year={2020}
}

@article{liu2026ssr,
  title={Ssr: Enhancing depth perception in vision-language models via rationale-guided spatial reasoning},
  author={Liu, Yang and Ma, Ming and Yu, Xiaomin and Ding, Pengxiang and Zhao, Han and Sun, Mingyang and Huang, Siteng and Wang, Donglin},
  journal={Advances in Neural Information Processing Systems},
  volume={38},
  pages={123926--123958},
  year={2026}
}

@inproceedings{liao2023slsd,
  title={Self-improvement of non-autoregressive model via sequence-level distillation},
  author={Liao, Yusheng and Jiang, Shuyang and Li, Yiqi and Wang, Yu and Wang, Yanfeng},
  booktitle={Proceedings of the 2023 conference on empirical methods in natural language processing},
  pages={14202--14212},
  year={2023}
}

@inproceedings{zhang2019your,
  title={Be your own teacher: Improve the performance of convolutional neural networks via self distillation},
  author={Zhang, Linfeng and Song, Jiebo and Gao, Anni and Chen, Jingwei and Bao, Chenglong and Ma, Kaisheng},
  booktitle={Proceedings of the IEEE/CVF international conference on computer vision},
  pages={3713--3722},
  year={2019}
}

@InProceedings{Su_2017_CVPR,
author = {Su, Bing and Hua, Gang},
title = {Order-Preserving Wasserstein Distance for Sequence Matching},
booktitle = {Proceedings of the IEEE Conference on Computer Vision and Pattern Recognition (CVPR)},
month = {July},
year = {2017}
}

@inproceedings{yang-etal-2018-modeling,
    title = "Modeling Localness for Self-Attention Networks",
    author = "Yang, Baosong  and
      Tu, Zhaopeng  and
      Wong, Derek F.  and
      Meng, Fandong  and
      Chao, Lidia S.  and
      Zhang, Tong",
    editor = "Riloff, Ellen  and
      Chiang, David  and
      Hockenmaier, Julia  and
      Tsujii, Jun{'}ichi",
    booktitle = "Proceedings of the 2018 Conference on Empirical Methods in Natural Language Processing",
    month = oct # "-" # nov,
    year = "2018",
    address = "Brussels, Belgium",
    publisher = "Association for Computational Linguistics",
    url = "https://aclanthology.org/D18-1475/",
    doi = "10.18653/v1/D18-1475",
    pages = "4449--4458"
}
